%% file: acl2020-MSC.tex
\documentclass[11pt,a4paper]{article}
\usepackage[hyperref]{acl2020}
\usepackage{times}
\usepackage{latexsym}

\usepackage{times}
\usepackage{latexsym}
\usepackage{graphicx}
\usepackage{amsmath}
\usepackage{booktabs}
\usepackage{algorithm}
\usepackage{algorithmic}
\usepackage{mathrsfs}
\usepackage{subfigure}
\usepackage{enumerate}
\usepackage{amsfonts}
\usepackage{multirow}
\usepackage{arydshln}
\usepackage{colortbl}
\usepackage{array}
\usepackage{url}
\usepackage{xcolor}
\usepackage{setspace}
\usepackage{pifont}
\usepackage{microtype}
\usepackage{tikz}
\newcommand*{\circled}[1]{\lower.7ex\hbox{\tikz\draw (0pt, 0pt)%
    circle (.5em) node {\makebox[1em][c]{\small #1}};}}
\aclfinalcopy 

\title{Multiscale Collaborative Deep Models for \\Neural Machine Translation}

\author{Xiangpeng Wei\textsuperscript{\rm 1,2}\Thanks{ Work done at Alibaba Group.},
	Heng Yu\textsuperscript{\rm 3}\Thanks{ Corresponding Author.},
	Yue Hu\textsuperscript{\rm 1,2$\dagger$}, 
	Yue Zhang\textsuperscript{\rm 4,5},
	Rongxiang Weng\textsuperscript{\rm 3},
	 Weihua Luo\textsuperscript{\rm 3}\\
	\textsuperscript{\rm 1}Institute of Information Engineering, Chinese Academy of Sciences, Beijing, China\\
	\textsuperscript{\rm 2}School of Cyber Security, University of Chinese Academy of Sciences, Beijing, China\\
	\textsuperscript{\rm 3}Machine Intelligence Technology Lab, Alibaba Group, Hangzhou, China\\
	\textsuperscript{\rm 4}School of Engineering, Westlake University, Hangzhou, China\\
	\textsuperscript{\rm 5}Institute of Advanced Technology, Westlake Institute for Advanced Study, Hangzhou, China\\
	\texttt{\{weixiangpeng,huyue\}@iie.ac.cn} \qquad \texttt{yue.zhang@wias.org.cn}\\
	\texttt{\{yuheng.yh,wengrx,weihua.luowh\}@alibaba-inc.com}\\
}

\date{}

\begin{document}
\maketitle
\begin{abstract}
\input{tex/abstract.tex}
\end{abstract}

\section{Introduction}
\input{tex/intro.tex}

\section{Background}
\input{tex/background.tex}

\section{Multiscale Collaborative Deep Model}
\input{tex/approach.tex}

\section{Experiments}
\input{tex/experiments.tex}

\section{Related Work}
\input{tex/related.tex}

\section{Conclusion and Future Work}
\input{tex/conclusion.tex}

\section*{Acknowledgments}
We would like to thank the anonymous reviewers for the
helpful comments. We also thank Xingxing Zhang, Luxi Xing and Kaixin Wu for their instructive suggestions and invaluable help. This work is supported by the National Key Research and Development Program (Grant No.2017YFB0803301).

\bibliography{acl2020}
\bibliographystyle{acl_natbib}

\appendix
\input{tex/appendix.tex}

\end{document}

%% file: tex/abstract.tex
Recent evidence reveals that Neural Machine Translation (NMT) models with deeper neural networks can be more effective but are difficult to train. In this paper, we present a \textbf{M}ulti\textbf{S}cale \textbf{C}ollaborative (\textsc{Msc}) framework to ease the training of NMT models that are substantially deeper than those used previously. We explicitly boost the gradient back-propagation from top to bottom levels by introducing a \textit{block-scale collaboration} mechanism into deep NMT models. Then, instead of forcing the whole encoder stack directly learns a desired representation, we let each encoder block learns a fine-grained representation and enhance it by encoding spatial dependencies using a \textit{context-scale collaboration}. We provide empirical evidence showing that the \textsc{Msc} nets are easy to optimize and can obtain improvements of translation quality from considerably increased depth. On IWSLT translation tasks with three translation directions, our extremely deep models (with 72-layer encoders) surpass strong baselines by +2.2$\sim$+3.1 BLEU points. In addition, our deep \textsc{Msc} achieves a BLEU score of 30.56 on WMT14 English$\rightarrow$German task that significantly outperforms state-of-the-art deep NMT models.

%% file: tex/intro.tex
Neural machine translation (NMT) directly models the entire translation process using a large neural network and has gained
rapid progress in recent years~\cite{Sutskever2014Seq2Seq,sennrich-etal-2016-neural}. The structure of NMT models has evolved quickly, such as RNN-based~\cite{Wu2016Google}, CNN-based~\cite{Gehring2017Convolutional} and attention-based~\cite{Vaswani2017Attention} systems. All of these models follow the encoder-decoder framework with attention~\cite{cho-etal-2014-learning,Bahdanau2014Neural,luong-etal-2015-effective} paradigm.

Deep neural networks have revolutionized the state-of-the-art in various communities, from computer vision to natural language processing. However, training deep neural networks has been always a challenging problem. To encourage gradient flow and error propagation, researchers in the field of computer vision have proposed various approaches, such as residual connections~\cite{he2016deep}, densely connected networks~\cite{huang2017densely} and deep layer aggregation~\cite{yu2018deep}. In natural language processing, constructing deep architectures has shown effectiveness in language modeling, question answering, text classification and natural language inference~\cite{peters-etal-2018-deep,radford2019language,al2019character,devlin-etal-2019-bert}. However, among existing NMT models, most of them are generally equipped with 4-8 encoder and decoder layers~\cite{Wu2016Google,Vaswani2017Attention}. Deep neural network has been explored relatively little in NMT.

Recent evidence~\cite{bapna-etal-2018-training,wang-etal-2019-learning} shows that model depth is indeed of importance to NMT, but a \textit{degradation} problem has been exposed: by simply stacking more layers, the translation quality gets saturated and then degrades rapidly. To address this problem, \citet{bapna-etal-2018-training} proposed a transparent attention mechanism to ease the optimization of the models with deeper encoders. \citet{wang-etal-2019-learning} continued this line of research but construct a much deeper encoder for Transformer by adopting the \textit{pre-norm} method that establishes a direct way to propagate error gradients from the top layer to bottom levels, and passing the combination of previous layers to the next. While notable gains have been reported over shallow models, the improvements of translation quality are limited when the model depth is beyond 20. In addition, degeneration of translation quality is still observed when the depth is beyond 30. As a result, two questions arise naturally: \textit{How to break the limitation of depth in NMT models?} and \textit{How to fully utilize the deeper structure to further improve the translation quality?}

In this paper, we address the degradation problem by proposing a \textbf{M}ulti\textbf{S}cale \textbf{C}ollaborative (\textsc{Msc}) framework for constructing NMT models with very deep encoders.\footnote{In our scenario, we mainly study the depth of encoders. The reason is similar in~\cite{wang-etal-2019-learning}: 1) encoders have a greater impact on performance than decoders; 2) increasing the depth of the decoder will significantly increase the complexity of inference.} In particular, the encoder and decoder of our model have the same number of \textit{blocks}, each consisting of one or several stacked layers.
Instead of relying on the whole encoder stack directly learns a desired representation, we let each encoder block learn a fine-grained representation and enhance it by encoding spatial dependences using a bottom-up network. For coordination, we attend each block of the decoder to both the corresponding representation of the encoder and the contextual representation with spatial dependences. This not only shortens the path of error propagation, but also helps to prevent the lower level information from being forgotten or diluted.

We conduct extensive experiments on WMT and IWSLT translation tasks, covering three translation directions with varying data conditions. On IWSLT translation tasks, we show that: 
\begin{itemize}
	\item While models with traditional stacking architecture exhibit worse performance on both training and validation data when depth increases, our framework is easy to optimize.
	\item The deep \textsc{Msc} nets (with 72-layer encoders) bring great improvements on translation quality from increased depth, producing results that substantially better than existing systems.
\end{itemize}

On the WMT14 English$\rightarrow$German task, we obtain improved results by deep \textsc{Msc} networks with a depth of 48 layers, outperforming strong baselines by +2.5 BLEU points, and also defeat state-of-the-art deep NMT models~\cite{wu-etal-2019-depth,zhangbiao-etal-2019-improving} with identical or less parameters.\footnote{\textsc{Msc} not only performs well on NMT but also is generalizable to other sequence-to-sequence generation tasks, such as abstractive summarization that is introduced in Appendix~\ref{Appendix-A}.}

\begin{figure*} 
	\centering 
	\subfigure[Overall framework.]{ 
		\label{fig:overview}
		\includegraphics[width=0.35\textwidth]{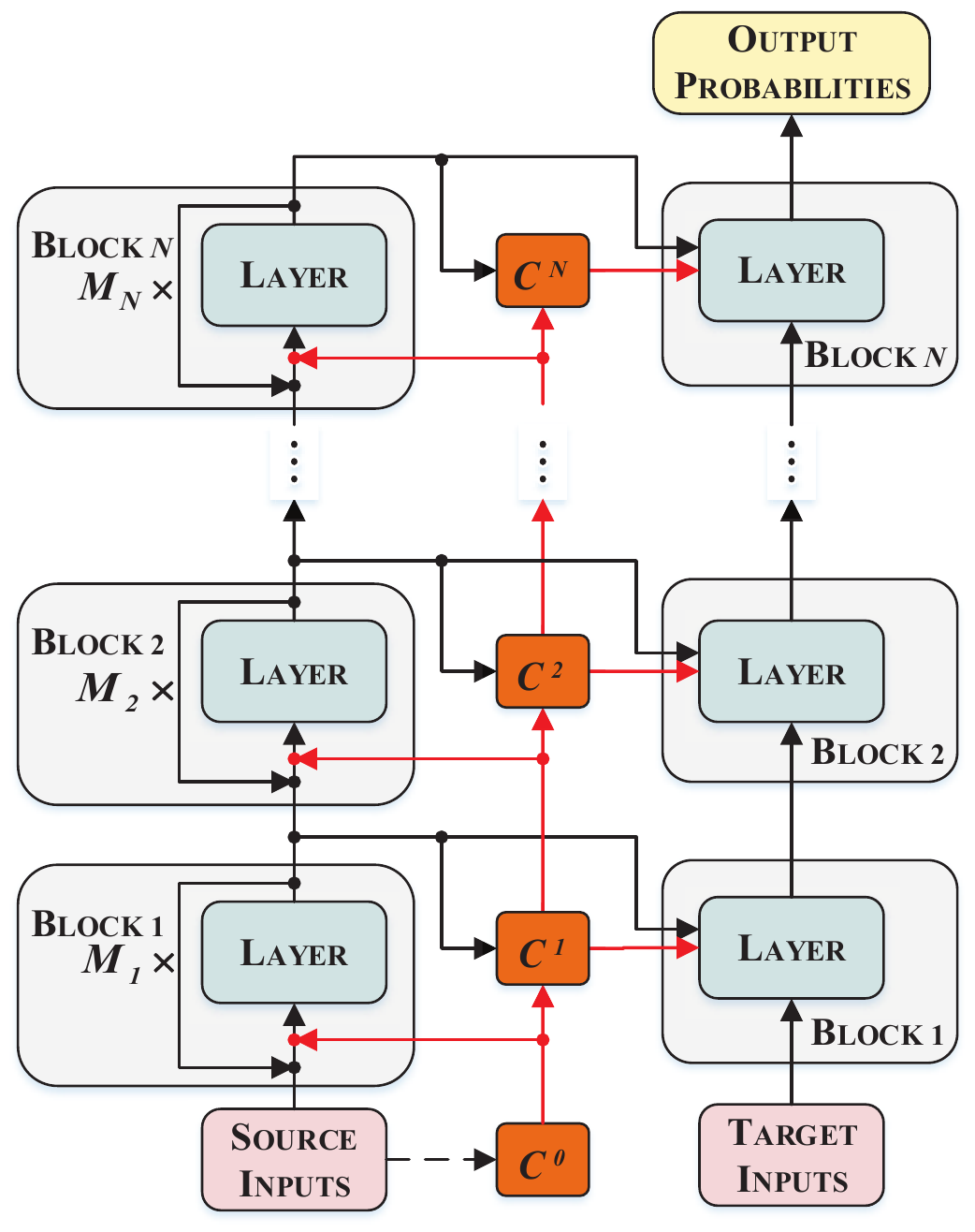}}
	\hspace{0.6cm}
	\subfigure[Detailed illustration of the $n$-th block.]{ 
		\label{fig:detail}
		\includegraphics[width=0.55\textwidth]{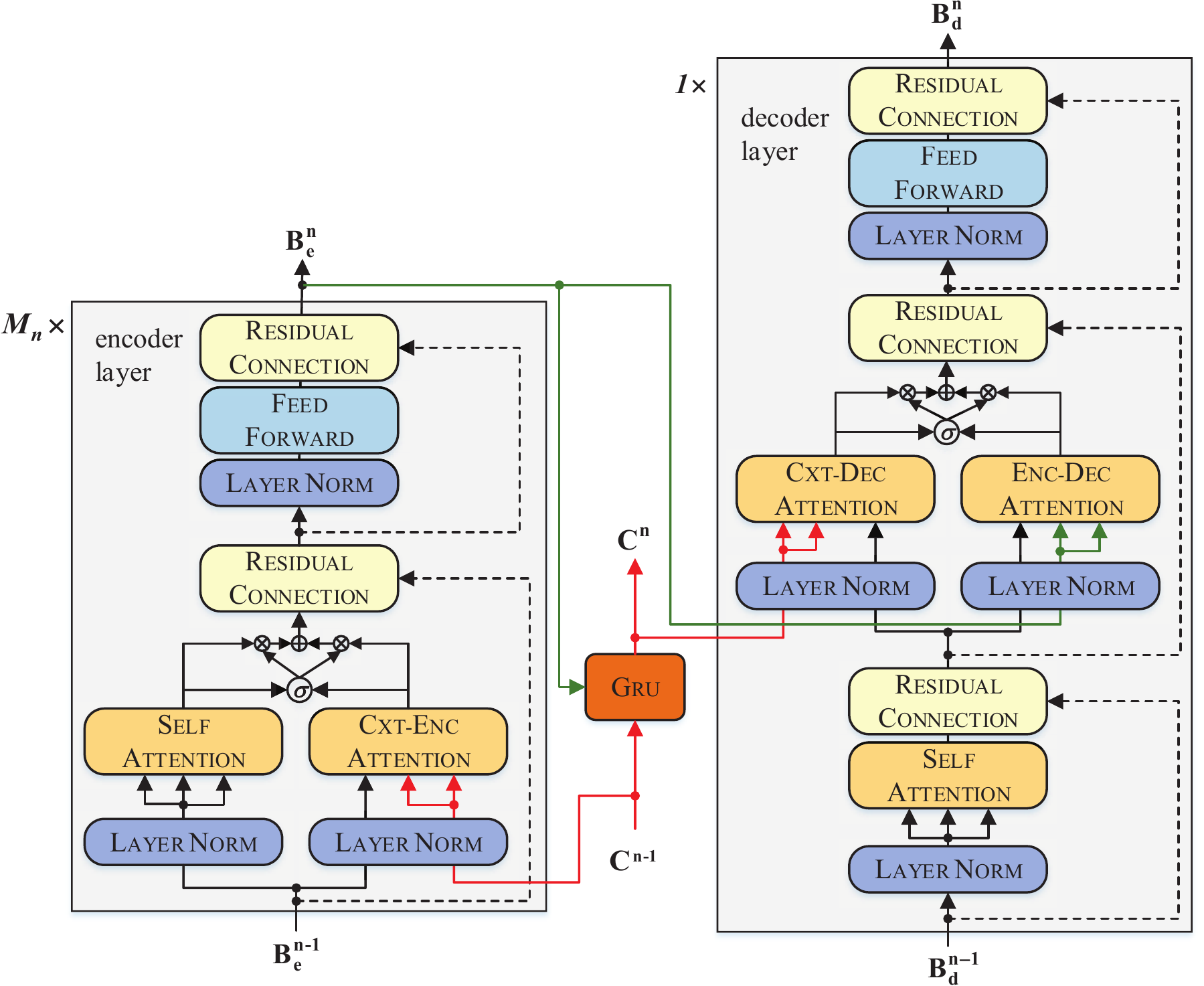}}
	\caption{\label{fig:model}Illustration of Multiscale Collaborative Deep NMT Model. $N$ is the number of encoder and decoder blocks. The $n$-th block of the encoder consists of $M_{n}$ layers, while each decoder block only contains one layer.}
\end{figure*}

%% file: tex/background.tex
Given a bilingual sentence pair $({\rm \bf x}, {\rm \bf y})$, an NMT model learns a set of parameters $\bf \Theta$ by maximizing the
log-likelihood $P({\rm \bf y}|{\rm \bf x}; {\bf \Theta})$, which is typically decomposed into the product of the conditional probability
of each target word: $P({\rm \bf y}|{\rm \bf x}; {\bf \Theta})=\prod_{t=1}^{T_{\rm y}}P({\rm \bf y}_t|{\rm \bf y}_{<t},{\rm \bf x}; {\bf \Theta})$, where $T_{\rm y}$ is the length of sentence ${\rm \bf y}$, ${\rm \bf y}_{<t}$ is the partial translation that contains the target tokens before position $t$. An encoder-decoder framework is commonly adopted to model the conditional probability $P({\rm \bf y}|{\rm \bf x}; {\bf \Theta})$, in which the encoder and decoder can be implemented as RNN~\cite{Wu2016Google}, CNN~\cite{Gehring2017Convolutional}, or Self-Attention network~\cite{Vaswani2017Attention}. Despite variant types of NMT architectures, multiple-layer encoder and decoder are generally employed to perform the translation task, and residual connections~\cite{he2016deep} are naturally introduced among layers, as ${\rm H}^{l} = \textsc{Layer}({\rm H}^{l-1};{\bf \Theta}^{l})+{\rm H}^{l-1}$, where ${\rm H}^{l}$ is the output of the $l$-th layer, $\text{{\sc Layer}}(\cdot)$ is the layer function and ${\bf \Theta}^{l}$ be the parameters.

We take the state-of-the-art Transformer as our baseline model. Specifically, the encoder consists of a stack of $L$ identical layers, each of which comprises two subcomponents: a self-attention mechanism followed by a feed-forward network. Layer normalization~\cite{ba2016layer} is applied to the input of each subcomponent (i.e., \textit{pre-norm}) and a residual skip connection~\cite{he2016deep} adds each subcomponent's input to its output. Formally,\begin{equation}
\begin{split}
{\rm O}_{e}^{l}&=\textsc{Attn}({\rm Q}_{e}^{l},{\rm K}_{e}^{l},{\rm V}_{e}^{l};{\bf \Theta}_{e}^{l}) + {\rm H}_{e}^{l-1},\\
{\rm H}_{e}^{l}&=\textsc{Ffn}(\textsc{Ln}({\rm O}_{e}^{l});{\bf \Theta}_{e}^{l}) + {\rm O}_{e}^{l},
\label{eq:encoder}
\end{split}
\end{equation}
where $\textsc{Ln}(\cdot)$, $\textsc{Attn}(\cdot)$ and $\textsc{Ffn}({\cdot})$ are layer normalization, attention mechanism, and feed-forward networks with ReLU activation in between, respectively. $\{{\rm Q}_{e}^{l},{\rm K}_{e}^{l},{\rm V}_{e}^{l}\}$ are query, key and value vectors that are transformed from the normalized ($l-1$)-th encoder layer $\textsc{Ln}({\rm H}_{e}^{l-1})$.

The decoder is similar in structure to the encoder except that it includes a standard attention mechanism after each self-attention network, which attends to the output of the encoder stack ${\rm H}_{e}^{L}$:
\begin{equation}
\begin{split}
{\rm O}_{d}^{l}&=\textsc{Attn}({\rm Q}_{d}^{l},{\rm K}_{d}^{l},{\rm V}_{d}^{l};{\bf \Theta}_{d}^{l}) + {\rm H}_{d}^{l-1},\\
{\rm S}_{d}^{l}&=\textsc{Attn}(\textsc{Ln}({\rm O}_{d}^{l}),{\rm K}_{e}^{L},{\rm V}_{e}^{L};{\bf \Theta}_{d}^{l}) + {\rm O}_{d}^{l},\\
{\rm H}_{d}^{l}&=\textsc{Ffn}(\textsc{Ln}({\rm S}_{d}^{l});{\bf \Theta}_{d}^{l}) + {\rm C}_{d}^{l},
\label{eq:decoder}
\end{split}
\end{equation}
where $\{{\rm Q}_{d}^{l},{\rm K}_{d}^{l},{\rm V}_{d}^{l}\}$ are transformed from the normalized ($l-1$)-th decoder layer $\textsc{Ln}({\rm H}_{d}^{l-1})$ and $\{{\rm K}_{e}^{L},{\rm V}_{e}^{L}\}$ are transformed from the top layer of the encoder. The top layer of the decoder ${\rm H}_{d}^{L}$ is used to generate the final output sequence. In the following sections, we simplify the equations as
\begin{equation}
\begin{split}
{\rm H}_{e}^{l}&=\mathcal{F}({\rm H}_{e}^{l-1};{\bf \Theta}_{e}^{l}) + {\rm H}_{e}^{l-1},\\
{\rm H}_{d}^{l}&=\mathcal{G}({\rm H}_{d}^{l-1},{\rm H}_{e}^{L};{\bf \Theta}_{d}^{l}) + {\rm H}_{d}^{l-1},
\end{split}
\end{equation}
for the encoder and decoder, respectively.

As discussed by~\citet{wang-etal-2019-learning}, applying layer normalization to the input of each subcomponent is the key to learning deep encoders, as it establishes a direct way to pass gradient from the top-most layer to bottom layers:
\begin{equation}
\frac{\partial \mathcal{L}}{\partial {\rm H}_{e}^{l}}=\frac{\partial \mathcal{L}}{\partial {\rm H}_{e}^{L}} \times (1+\sum_{j=l}^{L-1}\frac{\partial \mathcal{F}({\rm H}_{e}^{j};{\bf \Theta}_{e}^{j+1})}{\partial {\rm H}_{e}^{l}}),
\label{eq:pre-norm-grad}
\end{equation}
where $\mathcal{L}$ is the cross entropy loss. However, as pointed out by \citet{wang-etal-2019-learning} that it can be difficult to deepen the encoder for better translation quality. We argue that as the right-most term in Eq.~(\ref{eq:pre-norm-grad}) approaches 0 for the lower levels of the encoder, the parameters of which cannot be sufficiently trained using the error gradient $\frac{\partial \mathcal{L}}{\partial {\rm H}_{e}^{L}}$ only. To solve this problem, we propose a novel approach to shorten the path of error propagation from $\mathcal{L}$ to bottom layers of the encoder.

%% file: tex/approach.tex
In this section, we introduce the details of the proposed approach, a \textbf{M}ulti\textbf{S}cale \textbf{C}ollaborative (\textsc{Msc}) framework for constructing extremely deep NMT models. The framework of our method consists of two main components shown in Figure~\ref{fig:overview}. First, a \textit{block-scale collaboration} mechanism establishes shortcut connections from the lower levels of the encoder to the decoder (as described in~\ref{sec:bsc}), which is the key to training very deep NMT models. We give explanation by seeing the gradient propagation process. Second, we further enhance source representations with spatial dependencies by \textit{contextual collaboration}, which is discussed in Section~\ref{sec:csc}.

\subsection{Block-Scale Collaboration}
\label{sec:bsc}
An intuitive extension of naive stacking of layers is to group few stacked layers into a \textit{block}. We suppose that the encoder and decoder of our model have the same number of blocks (i.e., $N$). Each block of the encoder has $M_{n}$ ($n \in \{1,2,...,N\}$) identical layers, while each decoder block contains one layer. Thus, we can adjust the value of each $M_{n}$ flexibly to increase the depth of the encoder. Formally, for the $n$-th block of the encoder:
\begin{equation}
{\rm B}_{e}^{n}=\textsc{Block}_{e}({\rm B}_{e}^{n-1}),
\label{eq:encoder-block}
\end{equation}
where $\textsc{Block}_{e}(\cdot)$ is the block function, in which the layer function $\mathcal{F}(\cdot)$ is iterated $M_{n}$ times, i.e.
\begin{equation}
\begin{split}
{\rm B}_{e}^{n}&={\rm H}_{e}^{n, M_{n}},\\
{\rm H}_{e}^{n, l}&=\mathcal{F}({\rm H}_{e}^{n, l-1};{\bf \Theta}_{e}^{n,l}) + {\rm H}_{e}^{n, l-1}, \\
{\rm H}_{e}^{n, 0}&={\rm B}_{e}^{n-1},
\end{split}
\end{equation}
where $l \in \{1,2,...,M_{n}\}$, ${\rm H}_{e}^{n, l}$ and ${\bf \Theta}_{e}^{n,l}$ are the representation and parameters of the $l$-th layer in the $n$-th block, respectively. The decoder works in a similar way but the layer function $\mathcal{G}(\cdot)$ is iterated only once in each block,
\begin{equation}
\begin{split}
{\rm B}_{d}^{n}&=\textsc{Block}_{d}({\rm B}_{d}^{n-1}, {\rm B}_{e}^{n})\\
&=\mathcal{G}({\rm B}_{d}^{n-1}, {\rm B}_{e}^{n};{\bf \Theta}_{d}^{n}) + {\rm B}_{d}^{n-1}.
\label{eq:decoder-block}
\end{split}
\end{equation}
Each block of the decoder attends to the corresponding encoder block. \citet{he2018layer} proposed a model that learns the hidden representations in two corresponding encoder and decoder layers as the same semantic level through layer-wise coordination and parameter sharing. Inspired by this, we focus on efficiently training extremely deep NMT models through directly attending decoder to the lower-level layers of the encoder, rather than only to the final representation of the encoder stack.

The proposed block-scale collaboration (\textsc{Bsc}) mechanism can effectively boost gradient propagation from prediction loss to lower level encoder layers. For explaining this, see again Eq. (\ref{eq:pre-norm-grad}), which explains the error back-propagation of pre-norm Transformer. Formally, we let $\mathcal{L}$ be the prediction loss. The differential of $\mathcal{L}$ with respect to the $l$-th layer in the $n$-th block ${\rm H}_{e}^{n, l}$ can be calculated as:\footnote{For a detailed derivation, we refer the reader to Appendix~\ref{Appendix-B}.}
\begin{equation}
\begin{split}
&\frac{\partial \mathcal{L}}{\partial {\rm H}_{e}^{n, l}} \\
\!&=\!\frac{\partial \mathcal{L}}{\partial {\rm B}_{e}^{N}} \! \times \! \frac{\partial {\rm B}_{e}^{N}}{\partial {\rm H}_{e}^{n, l}} \!+\! \frac{\partial \mathcal{L}}{\partial {\rm B}_{e}^{n}} \! \times \! \frac{\partial {\rm B}_{e}^{n}}{\partial {\rm H}_{e}^{n, l}}\\
\!&=\! \underbrace{\frac{\partial \mathcal{L}}{\partial {\rm B}_{e}^{N}} \! \times \! (1 \!+\! \!\sum_{k=l+1}^{M_{n}}\! \frac{\partial {\rm H}_{e}^{n,k}}{\partial {\rm H}_{e}^{n,l}} \!+\! \sum_{i=n+1}^{N}\sum_{j=1}^{M_{i}}\frac{\partial {\rm H}_{e}^{i, j}}{\partial {\rm H}_{e}^{n, l}})}_{(a)}\\
\!&+\! \underbrace{\frac{\partial \mathcal{L}}{\partial {\rm B}_{e}^{n}} \! \times \! (1 \!+\! \sum_{k=l+1}^{M_{n}}\frac{\partial {\rm H}_{e}^{n,k}}{\partial {\rm H}_{e}^{n, l}})}_{(b)},
\end{split}
\label{eq:block-scale-grad}
\end{equation}
where term $(a)$ is equal to Eq. (\ref{eq:pre-norm-grad}). In addition to the straightforward path $\frac{\partial \mathcal{L}}{\partial {\rm B}_{e}^{N}}$ for parameter update from the top-most layer to lower ones, Eq. (\ref{eq:block-scale-grad}) also provides a complementary way to directly pass error gradient $\frac{\partial \mathcal{L}}{\partial {\rm B}_{e}^{n}}$ from top to bottom in the current block. Another benefit is that \textsc{Bsc} shortens the length of gradient pass chain (i.e., $M_{n} \ll L$).

\subsection{Contextual Collaboration}
\label{sec:csc}
To model long-term spatial dependencies and reuse global representations, we define a GRU~\cite{cho-etal-2014-learning} cell $\mathcal{Q}({\rm \bf c},\mathbf{\bar{\rm \bf x}})$, which maps a hidden state ${\rm \bf c}$ and an additional input $\mathbf{\bar{\rm \bf x}}$ into a new hidden state:
\begin{equation}
\begin{split}
{\rm C}^{n} &= \mathcal{Q}({\rm C}^{n-1},{\rm B}_{e}^{n}), n \in [1,N]\\
{\rm C}^{0} &= \mathcal{E}_{e},
\label{eq:Q}
\end{split}
\end{equation}
where $\mathcal{E}_{e}$ is the embedding matrix of the source input ${\rm \bf x}$. The new state ${\rm C}^{n}$ can be fused with each layer of the subsequent blocks in both the encoder and the decoder. Formally, ${\rm B}_{e}^{n}$ in Eq.(\ref{eq:encoder-block}) can be re-calculated in the following way:
\begin{equation}
\begin{split}
{\rm B}_{e}^{n}&={\rm H}_{e}^{n, M_{n}},\\
{\rm H}_{e}^{n, l}&=\mathcal{F}({\rm H}_{e}^{n, l-1},\textcolor{red}{{\rm C}^{n-1}};{\bf \Theta}_{e}^{n,l}) + {\rm H}_{e}^{n, l-1},\\
{\rm H}_{e}^{n, 0}&={\rm B}_{e}^{n-1}.
\label{eq:encoder-msc}
\end{split}
\end{equation}
Similarly, for decoder, we have
\begin{equation}
\begin{split}
{\rm B}_{d}^{n}&=\textsc{Block}_{d}({\rm B}_{d}^{n-1}, {\rm B}_{e}^{n})\\
&=\mathcal{G}({\rm B}_{d}^{n-1}, {\rm B}_{e}^{n},\textcolor{red}{{\rm C}^{n}};{\bf \Theta}_{d}^{n}) + {\rm B}_{d}^{n-1}.
\label{eq:decoder-msc}
\end{split}
\end{equation}

The above design is inspired by multiscale RNNs (\textsc{Mrnn})~\cite{schmidhuber1992learning,el1996hierarchical,koutnik2014clockwork,chung2016hierarchical}, which encode temporal dependencies with different timescales. Unlike \textsc{Mrnn}, our \textsc{Msc} enables each decoder block to attend to multi-granular source information with different space-scales, which helps to prevent the lower level information from being forgotten or diluted.

\paragraph{Feature Fusion:} We fuse the contextual representation with each layer of the encoder and decoder through attention. A detailed illustration of our algorithm is shown in Figure~\ref{fig:detail}. In particular, the $l$-th layer of the $n$-th encoder block $\mathcal{F}(\cdot;{\bf \Theta}_{e}^{n,l})$, $l \in [1,M_{n}]$ and $n \in [1,N]$,
\begin{equation}
\begin{split}
&{\rm O}_{e}^{n,l}\\
&=g_{e} \odot \textsc{Attn}_{h}({\rm H}_{e}^{n,l-1},{\rm H}_{e}^{n,l-1},{\rm H}_{e}^{n,l-1};{\bf \Theta}_{e}^{n,l})\\
&+\textcolor{red}{(1-g_{e})\odot\textsc{Attn}_{c}({\rm H}_{e}^{n,l-1},{\rm C}^{n-1},{\rm C}^{n-1};{\bf \Theta}_{e}^{n,l})}\\
&+{\rm H}_{e}^{n,l-1},\\
&g_{e} = \sigma(W_{1}\textsc{Attn}_{h}(\cdot)+W_{2}\textsc{Attn}_{c}(\cdot)+b),\\
\end{split}
\end{equation}
where $g_{e}$ is a gate unit, $\textsc{Attn}_{h}(\cdot)$ and $\textsc{Attn}_{c}(\cdot)$ are attention models (see Eq. (\ref{eq:encoder})) with different parameters. ${\rm O}_{e}^{n,l}$ is further processed by $\textsc{Ffn}(\cdot)$ to output the representation ${\rm H}_{e}^{n,l}$. Symmetrically, in the decoder, ${\rm S}_{d}^{n}$ in Eq. (\ref{eq:decoder}) can be calculated as
\begin{equation}
\begin{split}
{\rm S}_{d}^{n}&=g_{d} \odot \textsc{Attn}_{h}({\rm O}_{d}^{n},{\rm B}_{e}^{n},{\rm B}_{e}^{n};{\bf \Theta}_{d}^{n})\\
&+\textcolor{red}{(1-g_{d})\odot\textsc{Attn}_{c}({\rm O}_{d}^{n},{\rm C}^{n},{\rm C}^{n};{\bf \Theta}_{d}^{n})} \\
&+ {\rm O}_{d}^{l}
\end{split}
\end{equation}
where ${\rm O}_{d}^{n}$ is the output of the self-attention sub-layer defined in Eq. (\ref{eq:decoder}). $g_{d}$ is another gate unit.


%% file: tex/experiments.tex
We first evaluate the proposed method on IWSLT14 English$\leftrightarrow$German (En$\leftrightarrow$De) and IWSLT17 English$\rightarrow$French (En$\rightarrow$Fr) benchmarks. To make the results more convincing, we also experiment on a larger WMT14 English$\rightarrow$German (En$\rightarrow$De) dataset.

\subsection{Settings} 

\paragraph{Dataset.}
The dataset for IWSLT14 En$\leftrightarrow$De are as in~\citet{Ranzato2015Sequence}, with $160k$ sentence pairs for training and $7584$ sentence pairs for validation. The concatenated validation sets are used as the test set (dev2010, dev2012, tst2010, tst2011, tst2012). For En$\rightarrow$Fr, there are $236k$ sentence pairs for training and $10263$ for validation. The concatenated validation sets are used as the test set (dev2010, tst2010, tst2011, tst2012, tst2013, tst2014, tst2015). For all IWSLT translation tasks, we use a joint source and target vocabulary with $10k$ byte-pair-encoding (BPE) types~\cite{sennrich-etal-2016-neural}. For the WMT14 En$\rightarrow$De task, the training corpus is identical to previous work~\cite{Vaswani2017Attention,wang-etal-2019-learning}, which consists of about 4.5 million sentence pairs. All the data are tokenized using the script \texttt{tokenizer.pl} of Moses~\cite{koehn-etal-2007-moses} and segmented into subword symbols using jointly BPE with $32k$ merge operations. The shared source-target vocabulary contains about $37k$ BPE tokens. We use newstest2013 as the development set and newstest2014 as the test set. Following previous work, we evaluate IWSLT tasks with tokenized case-insensitive BLEU and report tokenized case-sensitive BLEU~\cite{papineni-etal-2002-bleu} for WMT14 En$\rightarrow$De.

\paragraph{Model Settings.} For IWSLT, the model configuration is \texttt{transformer\_iwslt}, representing a \texttt{small} model with embedding size $256$ and FFN layer dimension $512$. We train all models using the Adam optimizer ($\beta_1/\beta_2=0.9/0.98$) with adaptive learning rate schedule (warm-up step with 4K for shallow models, 8K for deep models) as in~\cite{Vaswani2017Attention} and label smoothing of $0.1$. Sentence pairs containing 16K$\sim$32K tokens are grouped into one batch. Unless otherwise stated, we train \texttt{small} models with 15K maximum steps, and decode sentences using beam search with a beam size of $5$ and length penalty of $1.0$.

For WMT14 En$\rightarrow$De, the model configuration is \texttt{transformer\_base/big}, with a embedding size of $512/1024$ and a FFN layer dimension of $2048/4096$. Experiments on WMT are conducted on 8 P100 GPUs. Following~\citet{ott-etal-2018-scaling}, we accumulate the gradient 8 iterations and then update to simulate a 64-GPU environment with a batch-size of 65K tokens per step. The Adam optimizer ($\beta_1/\beta_2=0.9/0.98$ for \texttt{base}, $\beta_1/\beta_2=0.9/0.998$) for \texttt{big}) and the warm-up strategy (8K steps for \texttt{base}, 16K steps for \texttt{big}) are also adopted. We use relatively larger batch size and dropout rate for deeper and bigger models for better convergence. The \texttt{transformer\_base/big} is updated for 100K/300K steps. For evaluation, we average the last 5/20 checkpoints for \texttt{base/big}, each of which is saved at the end of an epoch. Beam search is adopted with a width of $4$ and a length penalty of $0.6$. We use \texttt{multi-bleu.perl} to evaluate both IWSLT and WMT tasks for fair comparison with previous work.

\begin{table}[t!]
	\centering
		\begin{tabular}{p{2.1cm}|p{1.29cm}<\centering|p{1.29cm}<\centering|p{1.29cm}<\centering}
			\hline
			Depth & 36-layer & 54-layer & 72-layer\\
			\hline
			dec. ($N$) & 6 & 6 & 6 \\
			\hline
			enc. ($N \times M$) & 6$\times$6 & 6$\times$9 & 6$\times$12\\
			\hline
		\end{tabular}
	\caption{\label{table:msc-architecture}Deep architectures of \textsc{Msc} on IWSLT tasks. We simply set $M_{1}=\cdots={M}_{N}=M$.}
\end{table}

\begin{table*}[t!]
	\centering
	\begin{spacing}{1.1}
		\begin{tabular}{l|l|c|c|c|c|c}
			\hline
			\# & \textbf{Model} & \textbf{Param.} & \textbf{En$\rightarrow$De} & \textbf{De$\rightarrow$En} & \textbf{En$\rightarrow$Fr}&
			\textbf{$\Delta$Train}/\textbf{$\Delta$Dec}\\
			\hline
			1 &\texttt{small}, 6 layers, $dp_{a}=dp_{r}=0.1$ & 10.5M &27.23 & 32.73 & 41.19 & 17/1800 \\
			2 &w/ \textsc{Msc} & 15.6M & 27.49 & 33.10 & 41.53 & 17/1736 \\
			\hline
			3 &\textsc{Msc}, 36 layers, $dp_{a}=dp_{r}=0.3$ & 43.3M & 29.04 & 34.86 & 42.90 & 23/1498 \\
			4 & w/ 54 layers & 60.0M & 29.32 & 35.16 & 43.62 & 27/1412 \\
			5 & w/ 72 layers & 76.6M & - & - & - & - \\
			6 & w/ 72 layers, $\lambda_{l_2}=10^{-5}$ & 76.6M & \bf 29.67 & \bf 35.81 & \bf 44.15 & 30/1340 \\
			\hline
		\end{tabular}
	\end{spacing}
	\caption{\label{bleu-table-iwslt}BLEU scores [\%] of IWSLT translation tasks. \textbf{$\Delta$Train}/\textbf{$\Delta$Dec}: training time (hours)/decoding time (tokens per second) with a batch size of 32 and a beam size of 5. Dropout is applied to the residual connection ($dp_{r}$) and attention weights ($dp_{a}$). We apply L2 regularization to the weights of deeper encoders with $\lambda_{l_2}=10^{-5}$, which is only applied to the IWSLT tasks as the corpora are smaller and thus more regularization is required.}
\end{table*}

\subsection{Results}

We first evaluate 36-layer, 54-layer and 72-layer \textsc{Msc} nets on IWSLT tasks. Table~\ref{table:msc-architecture} summarizes the architecture. As shown in Table~\ref{bleu-table-iwslt}, applying \textsc{Msc} to the vanilla Transformer with 6 layers slightly increases translation quality by +0.26$\sim$+0.37 BLEU (\textcircled{\footnotesize 1}$\rightarrow$\textcircled{\footnotesize   2}). When the depth is increasing to 36, we use relatively larger dropout rate of $0.3$ and achieve substantially improvements (+1.4$\sim$+1.8 BLEU) over its shallow counterparts (\textcircled{\footnotesize 3} v.s. \textcircled{\footnotesize 2}). After that, we continue deepening the encoders in order, however, our extremely deep models (72 layers, \textcircled{\footnotesize 5}) suffer from \textit{overfitting} issue on the small IWSLT corpora, which cannot be solved by simply enlarging the dropout rate. We seek to solve this issue by applying L2 regularization to the weights of encoders with greatly increased depth. Results show that this works for deeper encoders (\textcircled{\footnotesize 6}).

\begin{table}[t!]
	\centering
	\begin{spacing}{1.1}
		\begin{tabular}{l|c}
			\hline
			 \textbf{Model} (\texttt{small}, 36 layers)&\textbf{BLEU}\\
			\hline
			\citet{bapna-etal-2018-training} & 28.09\\
			\citet{wang-etal-2019-learning} & 28.63\\
			\textsc{Msc} & 29.04\\
			\hline
			\textbf{Model} (\texttt{small}, 72 layers)&\\
			\hline
			\citet{bapna-etal-2018-training} & failed\\
			\citet{wang-etal-2019-learning} & 28.34\\
			\textsc{Msc} & 29.67\\
			\hline
		\end{tabular}
	\end{spacing}
	\caption{\label{bleu-table-iwslt-ende}Comparison with existing methods on IWSLT14 En$\rightarrow$De translation. For a fair comparison, we implemented all methods on the same Transformer backbone as well as model settings.}
\end{table}

We also report the inference speed in Table~\ref{bleu-table-iwslt} (the last column). As expected, the speed decreases with the depth of \textsc{Msc} increasing, which is consistent with observation of~\citet{wang-etal-2019-learning}. Compared to the baseline, \textsc{Msc} (72 layers) reduces decoding speed by 26\%. We leave further investigation on this issue to future work.

For fair comparisons, we implement existing methods~\cite{bapna-etal-2018-training,wang-etal-2019-learning} on the same vanilla Transformer backbone. We separately list the results of 36-layer and 72-layer encoders on the IWSLT14 En$\rightarrow$De task in Table~\ref{bleu-table-iwslt-ende}. The method of \citet{bapna-etal-2018-training} fail to train a very deep architecture while the method of \citet{wang-etal-2019-learning} is exposed a degradation phenomenon (28.63$\rightarrow$28.34). In contrast, \textsc{Msc} in both 36-layer and 72-layer cases outperform these methods. This suggests that our extremely deep models can easily bring improvements on translation quality from greatly increased depth, producing results substantially better than existing systems.

\begin{table}[t!]
	\centering
	\begin{spacing}{1.1}
		\begin{tabular}{l|r|c}
			\hline
			\textbf{Model} & \textbf{Param.} &\textbf{BLEU}\\
			\hline
			\citet{Vaswani2017Attention} & 213M & 28.4\\
			\citet{bapna-etal-2018-training} & 137M & 28.0\\
			\citet{dou-etal-2018-exploiting} &356M & 29.2\\
			\citet{he2018layer} & $^\ddagger$210M & 29.0\\
			\citet{wang-etal-2019-learning}& 137M & 29.3\\
			\citet{zhangbiao-etal-2019-improving}& 560M & 29.62\\
			\citet{wu-etal-2019-depth} & $^\ddagger$268M & \bf 29.9\\
			\hline
			{\sc Transformer} (\texttt{base}) &63M & 27.44\\
			\textsc{Msc}, 6 layers (\texttt{base}) & 73M & 27.68\\
			\textsc{Msc}, 36 layers (\texttt{base}) & 215M & 29.71\\
			\textsc{Msc}, 48 layers (\texttt{base}) & 272M & \bf 30.19\\
			\hline
			{\sc Transformer} (\texttt{big})& 211M & 28.86\\
			\textsc{Msc}, 6 layers (\texttt{big})& 286M & 29.17\\
			\textsc{Msc}, 18 layers (\texttt{big})& 512M & \bf 30.56\\
			\hline
		\end{tabular}
	\end{spacing}
	\caption{\label{bleu-table-wmt-ende}BLEU scores [\%] on WMT14 En$\rightarrow$De translation. $^{\ddagger}$ denotes an estimate value.}
\end{table}

Table~\ref{bleu-table-wmt-ende} lists the results on the WMT14 En$\rightarrow$De translation task and the comparison with the current state-of-the-art systems. The architectures ($N \times M$) of the 18-layer, 36-layer and 48-layer encoders are set as 6$\times$3, 6$\times$6 and 6$\times$8 respectively. We can see that incorporating our \textsc{Msc} into the shallow \texttt{base/big} contributes to +0.24/+0.31 BLEU (27.44$\rightarrow$27.68/28.86$\rightarrow$29.17) improvements under the same depth. When the depth grows, \textsc{Msc} demonstrates promising improvements of +1.39$\sim$+2.51 BLEU points over its shallow counterparts. It is worth noting that deep \textsc{Msc} with the \texttt{base} setting significantly outperforms the shallow one with the \texttt{big} setting (29.17$\rightarrow$30.19), though both of them have around the same number of parameters. Compared to existing models, our \textsc{Msc} outperforms the transparent model~\cite{bapna-etal-2018-training} (+2.2 BLEU) and the \textsc{Dlcl} model (+0.9 BLEU)~\cite{wang-etal-2019-learning}, two recent approaches for deep encoding. Compared to both the depth scaled model~\cite{zhangbiao-etal-2019-improving} and the current SOTA~\cite{wu-etal-2019-depth}, our \textsc{Msc} achieves better performance with identical or less parameters.

\subsection{Analysis}

\begin{figure} 
	\centering 
	\subfigure[Plain Network]{ 
		\label{fig:analysis:plain}
		\includegraphics[width=0.23\textwidth]{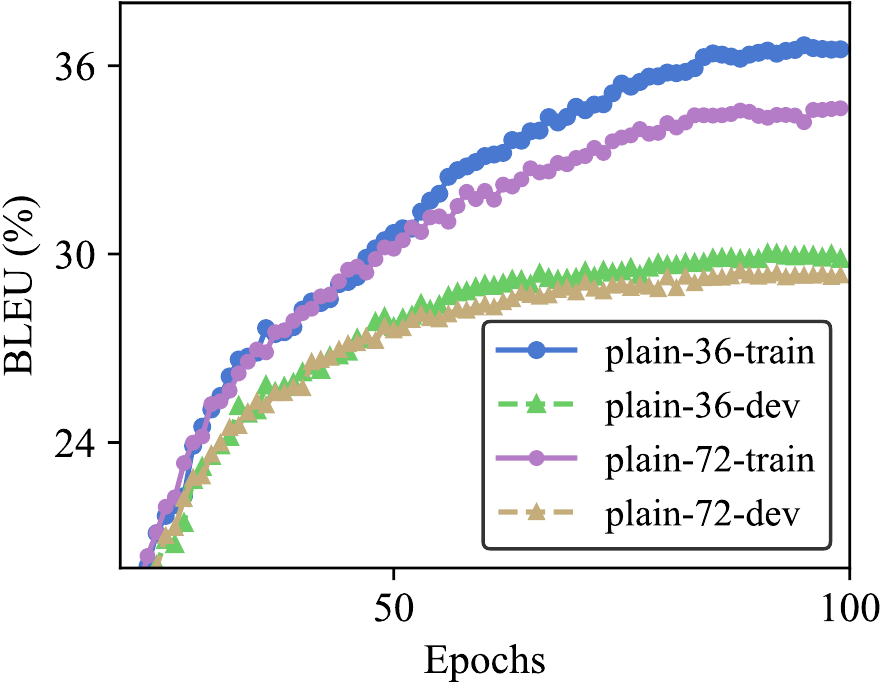}}
	\subfigure[\textsc{Msc} Network]{ 
		\label{fig:analysis:msc}
		\includegraphics[width=0.23\textwidth]{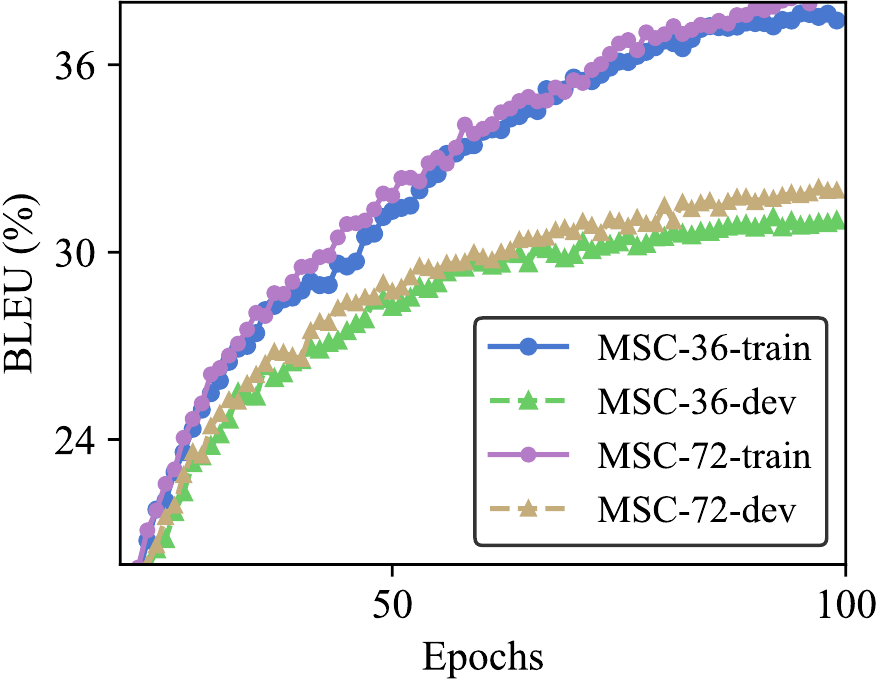}}
	\caption{\label{fig:msc}Illustration of the degradation problem on IWSLT14 En$\rightarrow$De task. We randomly select 3K sentence pairs from our training data for evaluation. For a fair comparison, we implemented all models on the same Transformer backbone as well as model settings.}
\end{figure}

\paragraph{Analysis of Degradation.}
We examine 36-layer and 72-layer \textit{plain} and \textsc{Msc} nets, respectively. For plain networks, we simply stack dozens of layers. As we can see from Figure~\ref{fig:analysis:plain}, the plain nets suffer from the degradation problem, which is not caused by overfitting, as they exhibit lower training BLEU. In contrast, the 72-layer \textsc{Msc} exhibits higher training BLEU than the 36-layer counterpart and is generalizable to the validation data. This indicates that our \textsc{Msc} can be more easily optimized with greatly increased depth.

\paragraph{Analysis of Handling Complicated Semantics.}
Although our \textsc{Msc} can enjoy improvements of BLEU score from increased depth, what does the models benefit from which is still implicit. To better understand this, we show the performance of deep \textsc{Msc} nets in handling sentences with complicated semantics. We assume that complicated sentences are difficult to fit with high prediction losses. Then we propose to use the modified prediction losses to identify these sentences:
\begin{equation}
\begin{split}
s({\rm \bf x},{\rm \bf y})=&\mathbb{E}\big[-\log P({\rm \bf y}|{\rm \bf x}; {\bf \Theta})\big]\\
&+ {\rm Std} \big[-\log P({\rm \bf y}|{\rm \bf x}; {\bf \Theta}) \big],
\end{split}
\end{equation}
where $\mathbb{E}\big[-\log P({\rm \bf y}|{\rm \bf x}; {\bf \Theta})\big]$ is approximated by:
\begin{equation}
\begin{split}
\mathbb{E}\big[-\log & P({\rm \bf y}|{\rm \bf x}; {\bf \Theta})\big] \\
&\approx \frac{1}{K} \sum_{k=1}^{K} \! -\log P({\rm \bf y}|{\rm \bf x}; {\bf \Theta}^{(k)}),
\end{split}
\end{equation}
where $\{{\bf \Theta}^{(k)}\}_{k=1}^{K}$ indicates model parameters for the last $K$ ($K\!=\!20$) checkpoints. ${\rm Std}[\cdot]$ is the standard deviation of prediction loss of sentence $\rm \bf y$ given sentence $\rm \bf x$, and the introduction of which aims to prevent training oscillations from affecting complicated sentences identification.

\begin{figure} 
	\centering 
	\includegraphics[width=0.44\textwidth]{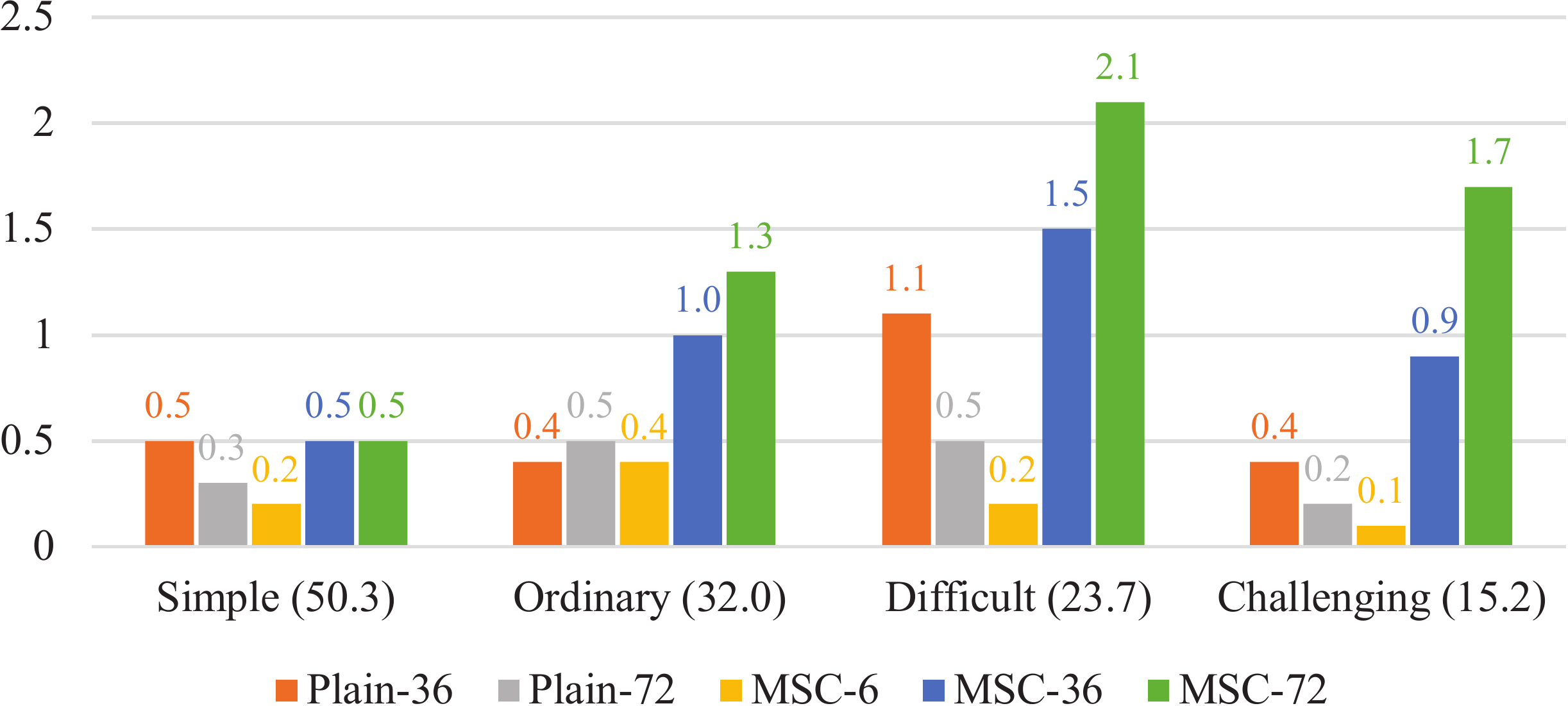}
	\caption{\label{fig:complicate}Comparison between plain nets and \textsc{Msc} nets on fine-grained test sets with increasing translation difficulty from ``Simple'' to ``Challenging''. Improvements (BLEU [\%]) of translation quality over the 6-layer plain net. Higher is better. The results of this baseline are enclosed in the parentheses.}
\end{figure}

We adopt a shallow plain net (\texttt{small}, 6 layers) to assign the prediction loss $s({\rm \bf x}, {\rm \bf y})$ to each sentence pair. Further, we split the IWSLT En$\rightarrow$De test set into 4 equal parts according to the prediction losses, which are pre-defined to have ``Simple'', ``Ordinary'', ``Difficult'' and ``Challenging'' translation difficulties, respectively.\footnote{The fine-grained test sets are publicly available at \url{https://github.com/pemywei/MSC-NMT/tree/master/IWSLT_En2De_Split_Test}.} Results on these fine-grained test sets are shown in Figure~\ref{fig:complicate}. First of all, all methods yield minor BLEU improvements over the baseline on the first sub-set that containing sentences with little difficulties to be translated. However, when the translation difficulty increases, the improvements of the deep \textsc{Msc} nets are expanded to around 2 BLEU. These results indicate that our MSC framework deals with sentences which are difficult to be translated well.

\begin{figure} 
	\centering 
	\subfigure[6-layer]{ 
		\label{fig:attn:shallow}
		\includegraphics[width=0.23\textwidth]{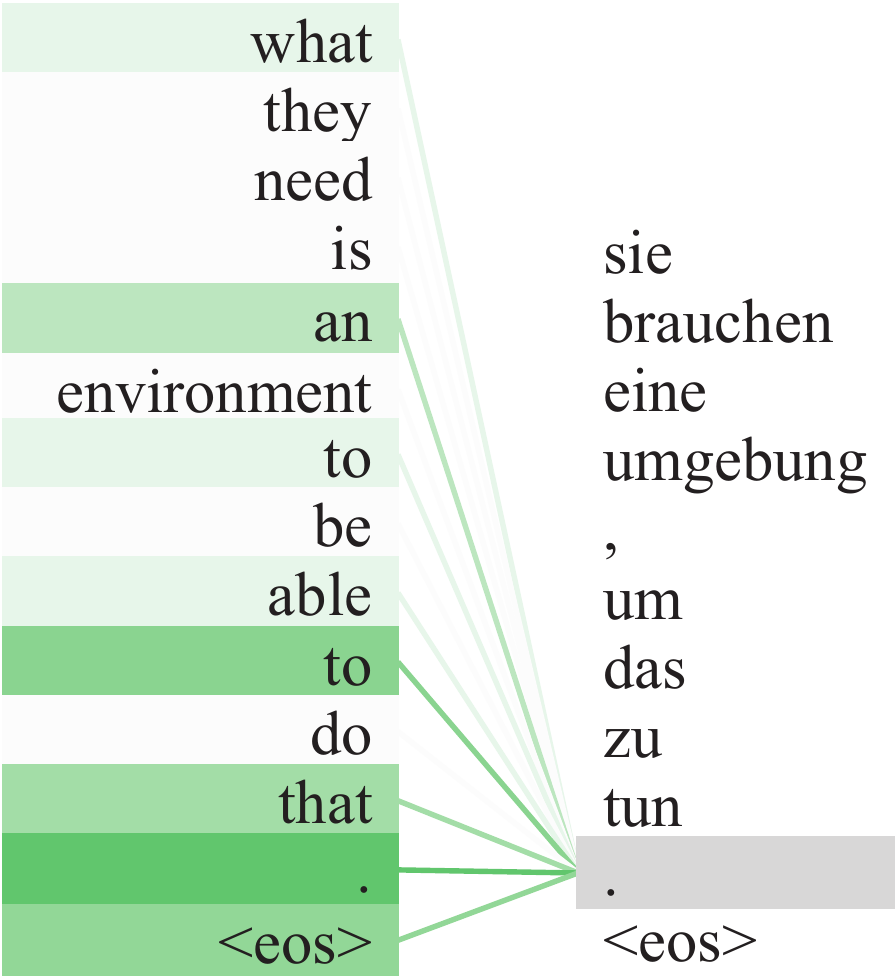}}
	\subfigure[72-layer]{ 
		\label{fig:attn:deep}
		\includegraphics[width=0.23\textwidth]{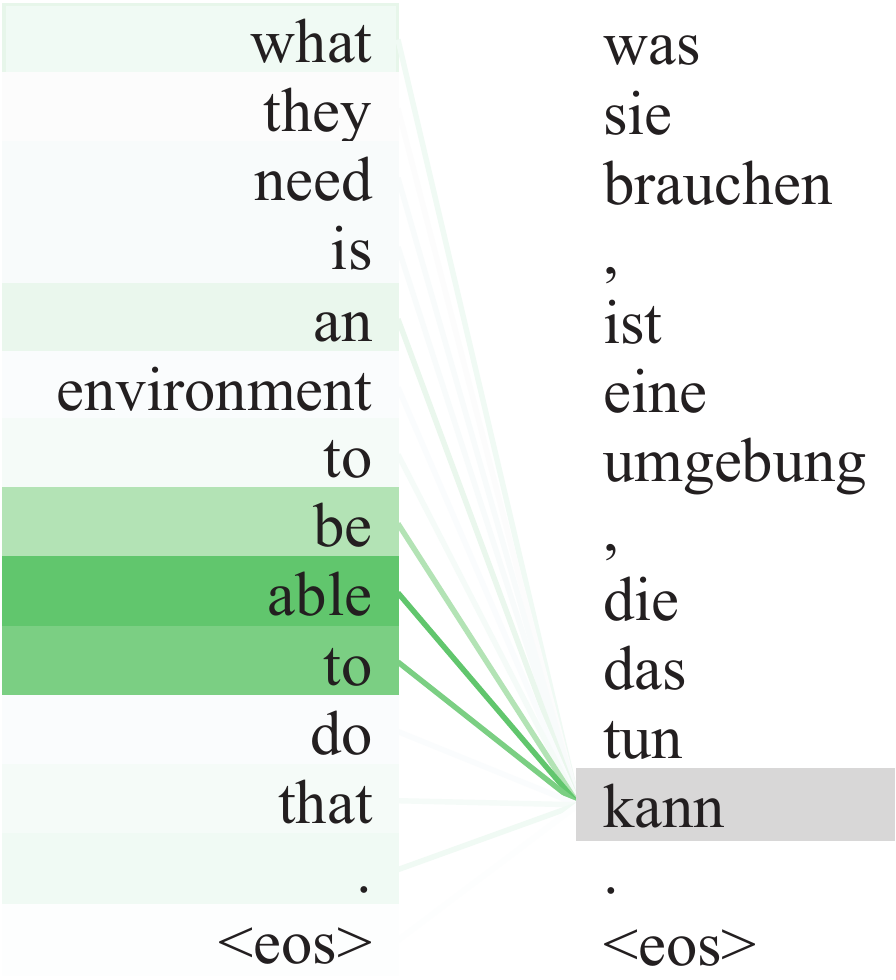}}
	\caption{\label{fig:visualization}Visualization of the attention weights from the top-most layer of the decoder for both shallow and deep \textsc{Msc} nets.}
\end{figure}

We also visualize the attention weights from the top-most layer of the decoder of both shallow and deep \textsc{Msc} nets in Figure~\ref{fig:visualization}. As shown in Figure~\ref{fig:attn:shallow}, when generating the next token of ``tun'', the shallow \textsc{Msc} attends to diverse tokens, such as ``to'', ``that'', ``.'' and ``eos'', which causes the generation of ``eos'' and the phrase ``be able to'' is mistakenly untranslated. Remarkably, the deep \textsc{Msc} (Figure~\ref{fig:attn:deep}) mostly focuses on the source tokens ``be'', ``able'' and ``to'', and translates this complicated sentence successfully. More cases can be found in Appendix~\ref{Appendix-C}. This kind of cases show the advantages of constructing extremely deep models for translating semantic-complicated sentences.

\begin{figure} 
	\centering 
	\includegraphics[width=0.48\textwidth]{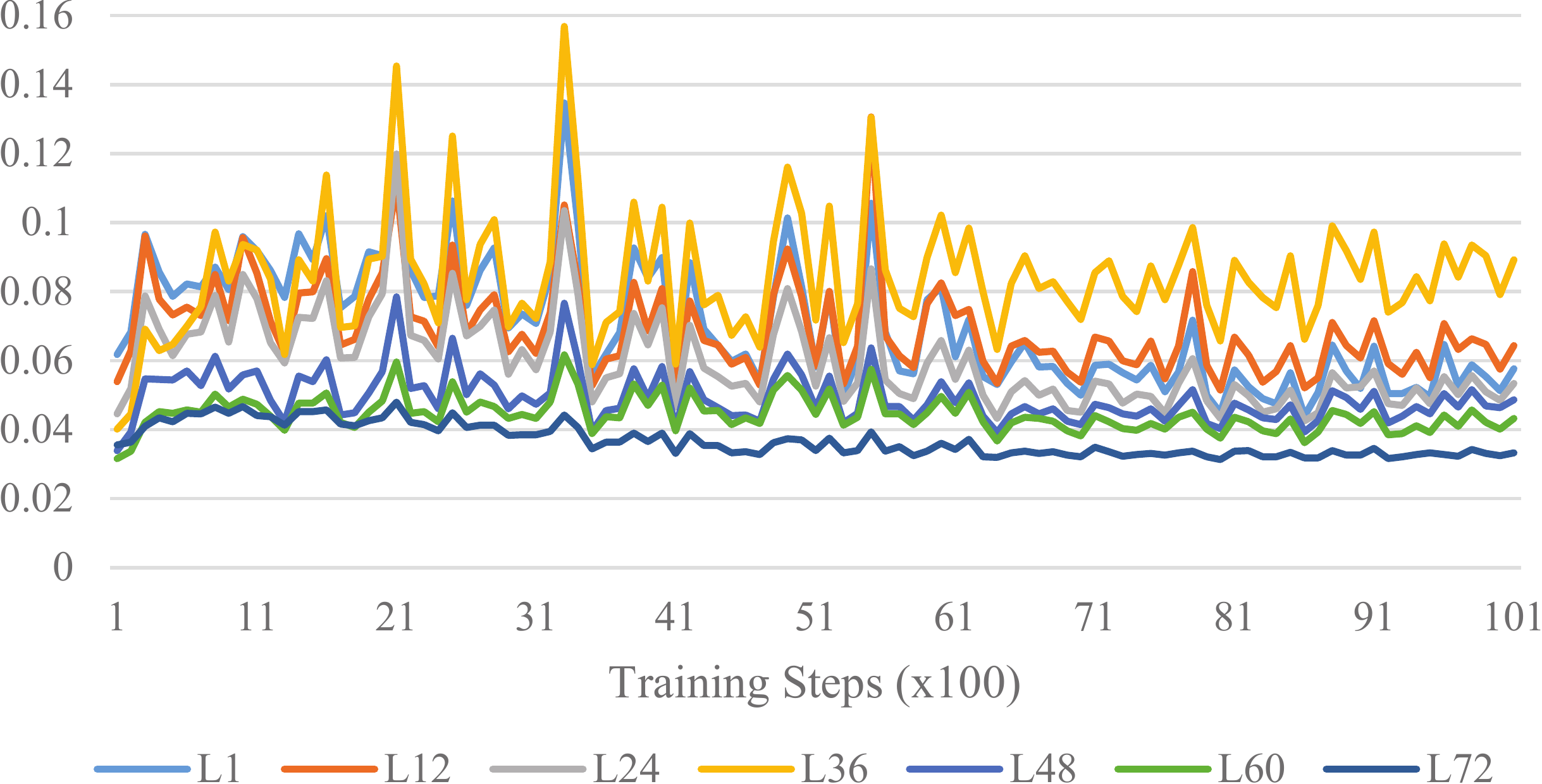}
	\caption{\label{fig:norm}Gradient norm (y-axis) of each encoder layer in 72-layer \textsc{Msc} over the fist 10k training steps. ``\emph{L$i$}'' denotes the $i$-th encoder layer. The \textsc{Msc} framework helps balance the gradient norm between top and bottom layers during training}
\end{figure}

\paragraph{Analysis of Error Propagation.}
To understand the propagation process of training signals, we collect the gradient norm of each encoder layer during training. Results in Figure~\ref{fig:norm} show that with the \textsc{Msc} framework each layer enjoys a certain value of gradient for parameter update, and the error signals traverse along the depth of the model without hindrance. \textsc{Msc} helps balance the gradient norm between top and bottom layers in deep models.

\paragraph{Ablation Study.}
We conduct ablation study to investigate the performance of each component of our model. The results are reported in Table~\ref{table:ablation}: (1) We use simple element-wise addition for feature fusion instead of using a gated combination as introduced in Section~\ref{sec:csc}. This method achieves a 29.45 BLEU, which is lower than the best result. We additionally modify the implementation of the contextual collaboration cell $\mathcal{Q}(\cdot)$ as $\textsc{Ffn}(\cdot)$, which shows that the performance is reduced by 0.5 BLEU. (2) Removing \textsc{Cxt-Enc Attention} and/or contextual collaboration makes the BLEU score drop by $\sim$0.7, which suggests that multiscale collaboration helps in constructing extremely deep models. (3) Considering that the deep \textsc{Msc} introduces more parameters, we also train another two \textsc{Msc} models with about the same or double number of parameters: with 18/36 layers, embedding size 512 and \textsc{Ffn} layer dimension 1024. These models underperform the deeper 72-layer model, which shows that the number of parameters is not the key to the improvement.

\begin{table}[t!]
	\begin{center}
		\begin{tabular}{p{5.9cm}|p{1.cm}<\centering}
			\hline
			\textbf{Model} & \textbf{BLEU}\\
			\hline
			\textsc{Msc}, 72 layers & 29.67\\
			\hline
			- feature fusion with addition & 29.45\\
			- implement $\mathcal{Q}(\cdot)$ in Eq. (\ref{eq:Q}) as $\textsc{Ffn}(\cdot)$ & 29.17\\
			\hline
			- remove \textsc{Cxt-Enc Attention} & 28.99\\
			- remove contextual collaboration & 28.94\\
			\hline
			\textsc{Msc}, 18 layers (emb=512, ffn=1024) & 29.08 \\
			\textsc{Msc}, 36 layers (emb=512, ffn=1024) & 29.41 \\
			\hline
		\end{tabular}
	\end{center}
    \caption{\label{table:ablation} Ablation study on IWSLT14 En$\rightarrow$De task.}
\end{table}

%% file: tex/related.tex
Researchers have constructed deep NMT models that use linear connections to reduce the gradient propagation length inside the topology~\cite{zhou-etal-2016-deep,wang-etal-2017-deep,ZhangDeepAttn2018} or read-write operations on stacked layers of memories~\cite{meng2015deep}. Such work has been conducted on the basis of the conventional RNN architectures and may not be fully applicable to the advanced Transformer.

Recently, \citet{bapna-etal-2018-training} introduced a transparent network into NMT models to ease the optimization of models with deeper encoders. To improve gradient flow they let each decoder layer find an unique weighted combination of all encoder layer outputs, instead of just the top encoder layer. \citet{wang-etal-2019-learning} found that adopting the proper use of layer normalization helps to learn deep encoders. A method was further proposed to combine layers and encourage gradient flow by simple shortcut connections. \citet{zhangbiao-etal-2019-improving} introduced a depth-scaled initialization to improve norm preservation and proposed a merged attention sublayer to avoid the computational overhead for deep models. Researchers have also explored growing NMT models in two stages~\cite{wu-etal-2019-depth}, in which shallow encoders and decoders are trained in the first stage and subsequently held constant, when another set of shallow layers are stacked on the top. In concurrent work, \citet{Xu2019Why} studied the effect of the computation order of residual connection and layer normalization, and proposed an parameter initialization method with Lipschitz restrictions to ensure the convergence of deep Transformers. Our method significantly differs from these methods, solving the problem by associating the decoder with the encoder with multi-granular dependencies in different space-scales.

Exploiting deep representations have been studied to strengthen feature propagation and encourage feature reuse in NMT~\cite{shen-etal-2018-dense,dou-etal-2018-exploiting,dou2019dynamic,wang-etal-2019-exploiting}. All of these works mainly attend the decoder to the final output of the encoder stack, we instead coordinate the encoder and the decoder at earlier stage.

%% file: tex/conclusion.tex
In this paper, we propose a multisacle collaborative framework to ease the training of extremely deep NMT models. Specifically, instead of the top-most representation of the encoder stack, we attend the decoder to multi-granular source information with different space-scales. We have shown that the proposed approach boosts the training of very deep models and can bring improvements on translation quality from greatly increased depth. Experiments on various language pairs show that the \textsc{Msc} achieves prominent improvements over strong baselines as well as previous deep models.

In the future, we would like to extend our model to extremely large datasets, such as WMT'14 English-to-French with about 36M sentence-pairs. And the deeper \textsc{Msc} model results in high computational overhead, to address this issue, we would like to apply the average attention network~\cite{zhang-etal-2018-accelerating} to our deep \textsc{Msc} models.

%% file: tex/appendix.tex
\section{Abstractive Summarization}
\label{Appendix-A}
We further verify the effectiveness of \textsc{Msc} on text summarization. Automatic text summarization produces a concise and fluent summary conveying the key information in the input (e.g., a news article). We focus on abstractive summarization, a generation task aims to generate the summary of a document with rewriting. We use the non-anonymized version of the CNN/DailyMail (CNNDM) dataset~\cite{see-etal-2017-get} for evaluation. We preprocessed the dataset using the scripts from the authors of~\citet{see-etal-2017-get},\footnote{\url{https://github.com/abisee/cnn-dailymail}} and the resulting dataset contains 287,226 documents with summaries for training, 13,368 for validation and 11,490 for test.

We still adopt the Transformer~\cite{Vaswani2017Attention} as our backbone, with a embedding size of 512 and FFN layer dimension of 1024. We train our model on the training set for 30 epochs, and also use label smoothing with rate of 0.1. We set batch size to 32, and maximum length to 768. During decoding, we use beam search with beam size of 5. The input document is truncated to the first 640 tokens. We remove duplicated trigrams in beam search, and tweak the maximum summary length on the development set~\cite{Paulus2017A,edunov-etal-2019-pre}. We use the F1 version of ROUGE~\cite{lin2004rouge} as the evaluation metric.

\begin{table}[t!]
	\centering
	\begin{spacing}{1.1}
		\begin{tabular}{l|c|c|c}
			\hline
			\bf Model & \bf R-1 & \bf R-2 & \bf R-L\\
			\hline
			\multicolumn{4}{c}{\textit{Extractive Summarization}}\\
			\hline
			Lead3 & 40.38 & 17.61 & 36.59\\
			\textsc{Hibert}$_{M}$& 42.37 & 19.95 & 38.83\\
			\citet{liu2019finetune}& 43.25 & 20.24 & 39.63\\
			\hline
			\multicolumn{4}{c}{\textit{Abstractive Summarization}}\\
			\hline
			PGNet & 39.53 & 17.28 & 36.38\\
			Bottom-Up& 41.22 & 18.68 & 38.34\\
			S2S-ELMo & 41.56 & 18.94 & 38.47\\
			\hline
			\textsc{Transformer} & 40.28 & 17.87 & 37.25\\
			\textsc{HierTrans} (36 L)& 41.22 & 18.97 & 38.45\\
			\textsc{Msc} (36 L) & \bf 41.96 & \bf 19.50 & \bf 39.07\\
			\hline
		\end{tabular}
	\end{spacing}
	\caption{\label{rouge-table-CNNDM}Results on CNNDM summarization using ROUGE-1 (R-1), ROUGE-2 (R-2), and ROUGE-L (R-L). ``36L'' is short for ``36-layer encoder''.}
\end{table}

In Table~\ref{rouge-table-CNNDM}, we compare \textsc{Msc} (36 layers) against the baseline and several state-of-the-art models on CNN/DailyMail, with extractive models in the top block and abstractive models in the bottom block. Lead3 is a baseline which simply selects the first three sentences as the summary. \textsc{Hibert}$_{M}$~\cite{zhang-etal-2019-hibert} adds the large open-domain unlabeled data to pre-train hierarchical transformer encoders and fine-tune on the extractive summarization task. We also include in Table~\ref{rouge-table-CNNDM} the best reported extractive summarization result taken from~\cite{liu2019finetune} on the dataset. PGNet~\cite{see-etal-2017-get}, Bottom-Up~\cite{gehrmann-etal-2018-bottom} and S2S-ELMo~\cite{edunov-etal-2019-pre} are all sequence to sequence learning based models with copy and coverage modeling, bottom-up content selecting and pre-trained ELMo representations augmenting. We also implemented two baselines. One is the standard 6-layer Transformer model. We can see that the deep \textsc{Msc} leads to a +1.8 ROUGE improvement over \textsc{Transformer}. The other baseline is the hierarchical transformer summarization model (\textsc{HierTrans}), which involevs both a sentence-level and a document-level transformer encoders, as well as a standard transformer decoder. Note the settings for both encoders are the same (each of them have $L$=18, emb=512, ffn=1024, head=8). The deep \textsc{Msc} outperforms \textsc{HierTrans} by 0.5 to 0.7 ROUGE with the same depth of encoders.

\section{Derivations of Block-Scale Collaboration}
\label{Appendix-B}
In pre-norm Transformer, a general transformation can be formulated as:
\begin{equation}
	\begin{split}
		{\rm H}^{l}&=\mathcal{F}({\rm H}^{l-1};{\bf \Theta}^{l}) + {\rm H}^{l-1},
	\end{split}
	\label{eq:transformation}
\end{equation}
where ${\rm H}^{l-1}$ and ${\rm H}^{l}$ are the input and output of the $l$-th layer. For the Block-Scale Collaboration framework, there are two channels for passing error gradients from the prediction loss $\mathcal{L}$ to encoder layers, which are from the top-most layers of the whole encoder stack ${\rm H}_{e}^{N, M_{N}}$ (identical to ${\rm B}_{e}^{N}$) and the current bock ${\rm H}_{e}^{n, M_{n}}$ (identical to ${\rm B}_{e}^{n}$), respectively. From the chain rule of back propagation we can obtain:
\begin{equation}
	\frac{\partial \mathcal{L}}{\partial {\rm H}_{e}^{n, l}}=\frac{\partial \mathcal{L}}{\partial {\rm B}_{e}^{N}} \times \frac{\partial {\rm B}_{e}^{N}}{\partial {\rm H}_{e}^{n, l}} + \frac{\partial \mathcal{L}}{\partial {\rm B}_{e}^{n}} \times \frac{\partial {\rm B}_{e}^{n}}{\partial {\rm H}_{e}^{n, l}}.
	\label{eq:chain-rule}
\end{equation}
We can recursively use Eq. (\ref{eq:transformation}) to formulate that
\begin{equation}
	\begin{split}
		{\rm B}_{e}^{N} =& {\rm H}_{e}^{N, M_{N}}\\
		=&{\rm H}_{e}^{n, l} + \sum_{k=l+1}^{M_{n}} \mathcal{F}({\rm H}_{e}^{n,k-1};{\bf \Theta}_{e}^{n,k})\\
		&+ \sum_{i=n+1}^{N}\sum_{j=1}^{M_{i}}\mathcal{F}({\rm H}_{e}^{i,j-1};{\bf \Theta}_{e}^{i,j}),
	\end{split}
\end{equation}
and
\begin{equation}
	\begin{split}
		{\rm B}_{e}^{N} =&{\rm H}_{e}^{n, M_{n}}\\
		=&{\rm H}_{e}^{n, l} + \sum_{k=l+1}^{M_{n}} \mathcal{F}({\rm H}_{e}^{n,k-1};{\bf \Theta}_{e}^{n,k}),
	\end{split}
\end{equation}
respectively. In this way, the derivations of ${\rm B}_{e}^{N}$ and ${\rm B}_{e}^{n}$ with respect to ${\rm H}_{e}^{n, l}$ can be calculated as:
\begin{equation}
	\begin{split}
		\frac{\partial {\rm B}_{e}^{N}}{\partial {\rm H}_{e}^{n,l}} = 1 &+ \sum_{k=l+1}^{M_{n}}\! \frac{\partial \mathcal{F}({\rm H}_{e}^{n,k-1};{\bf \Theta}_{e}^{n,k})}{\partial {\rm H}_{e}^{n,l}}\\
		&+\sum_{i=n+1}^{N}\sum_{j=1}^{M_{i}}\frac{\partial \mathcal{F}({\rm H}_{e}^{i,j-1};{\bf \Theta}_{e}^{i,j})}{\partial {\rm H}_{e}^{n, l}},\\
		\frac{\partial {\rm B}_{e}^{n}}{\partial {\rm H}_{e}^{n,l}} = 1 &+ \sum_{k=l+1}^{M_{n}}\frac{\partial \mathcal{F}({\rm H}_{e}^{n,k-1};{\bf \Theta}_{e}^{n,k})}{\partial {\rm H}_{e}^{n, l}}.
	\end{split}
	\label{eq:sub-item}
\end{equation}
Finally, we can put Eq. (\ref{eq:sub-item}) into Eq. (\ref{eq:chain-rule}) and obtain:
\begin{equation}
	\begin{split}
		&\frac{\partial \mathcal{L}}{\partial {\rm H}_{e}^{n, l}} \\
		\!&=\!\frac{\partial \mathcal{L}}{\partial {\rm B}_{e}^{N}} \! \times \! \frac{\partial {\rm B}_{e}^{N}}{\partial {\rm H}_{e}^{n, l}} \!+\! \frac{\partial \mathcal{L}}{\partial {\rm B}_{e}^{n}} \! \times \! \frac{\partial {\rm B}_{e}^{n}}{\partial {\rm H}_{e}^{n, l}}\\
		\!&=\! \underbrace{\frac{\partial \mathcal{L}}{\partial {\rm B}_{e}^{N}} \! \times \! (1 \!+\! \!\sum_{k=l+1}^{M_{n}}\! \frac{\partial {\rm H}_{e}^{n,k}}{\partial {\rm H}_{e}^{n,l}} \!+\! \sum_{i=n+1}^{N}\sum_{j=1}^{M_{i}}\frac{\partial {\rm H}_{e}^{i, j}}{\partial {\rm H}_{e}^{n, l}})}_{(a)}\\
		\!&+\! \underbrace{\frac{\partial \mathcal{L}}{\partial {\rm B}_{e}^{n}} \! \times \! (1 \!+\! \sum_{k=l+1}^{M_{n}}\frac{\partial {\rm H}_{e}^{n,k}}{\partial {\rm H}_{e}^{n, l}})}_{(b)},
	\end{split}
	\label{eq:block-scale-grad-detail}
\end{equation}

\begin{figure} 
	\centering 
	\subfigure[Case 1: 6-layer]{ 
		\label{fig:attn:shallow-1}
		\includegraphics[width=0.2\textwidth]{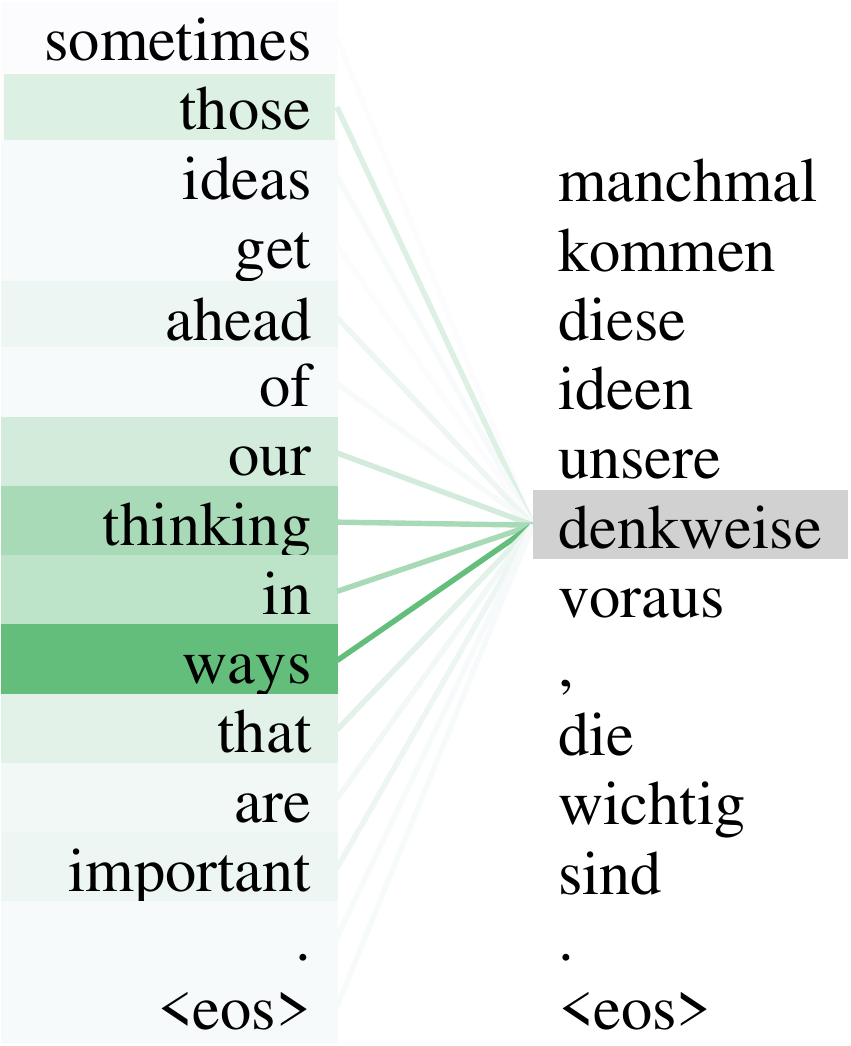}}
	\hspace{0.4cm}
	\subfigure[Case 1: 72-layer]{ 
		\label{fig:attn:deep-1}
		\includegraphics[width=0.2\textwidth]{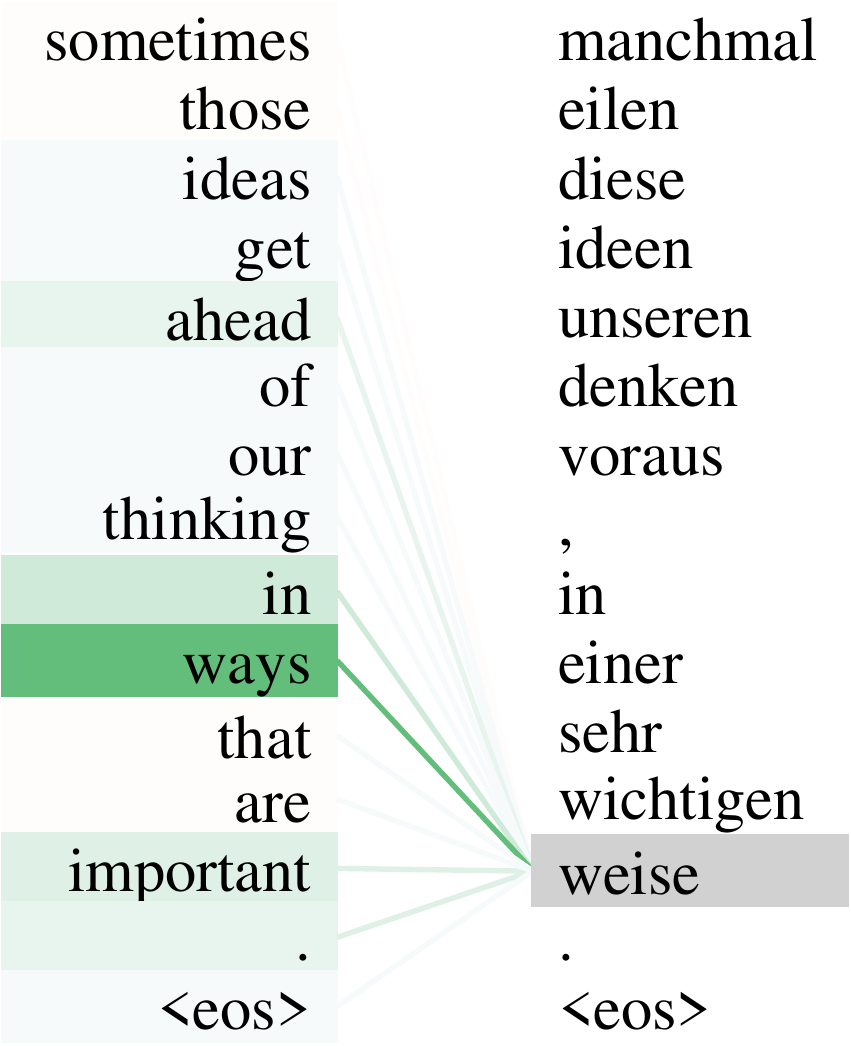}}
	
	\subfigure[Case 2: 6-layer]{ 
		\label{fig:attn:shallow-2}
		\includegraphics[width=0.2\textwidth]{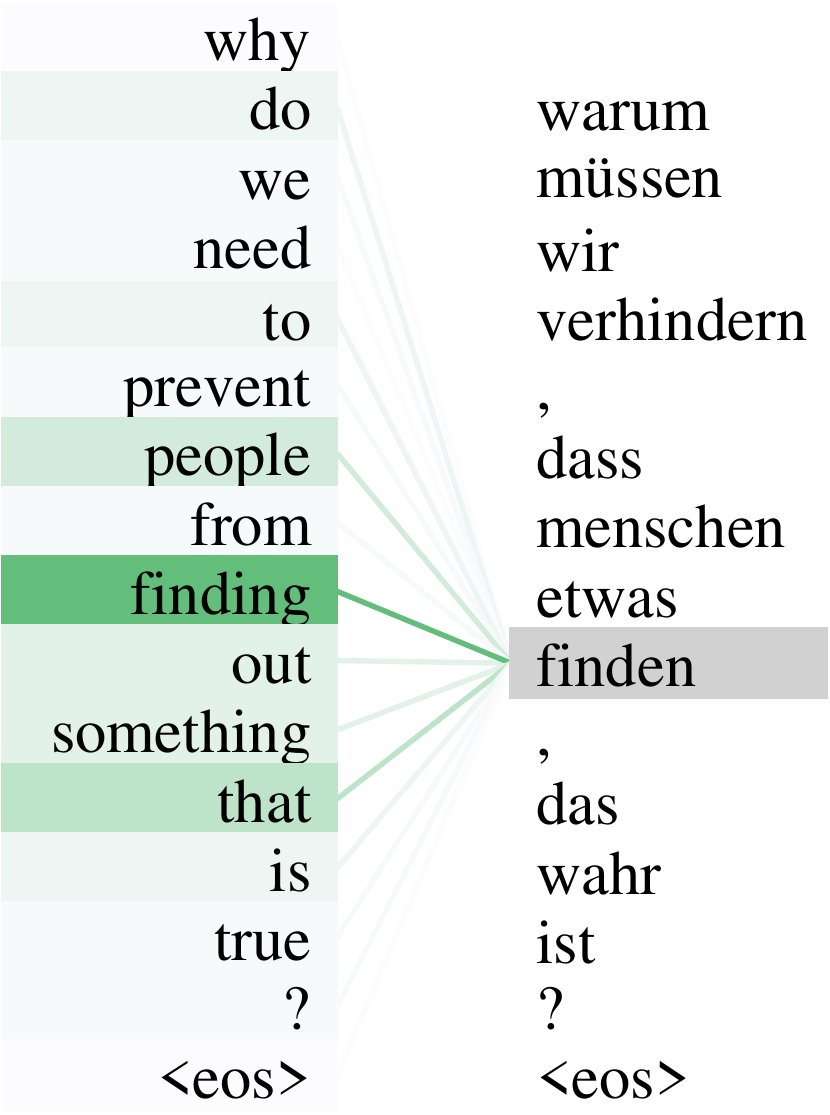}}
	\hspace{0.4cm}
	\subfigure[Case 2: 72-layer]{ 
		\label{fig:attn:deep-2}
		\includegraphics[width=0.2\textwidth]{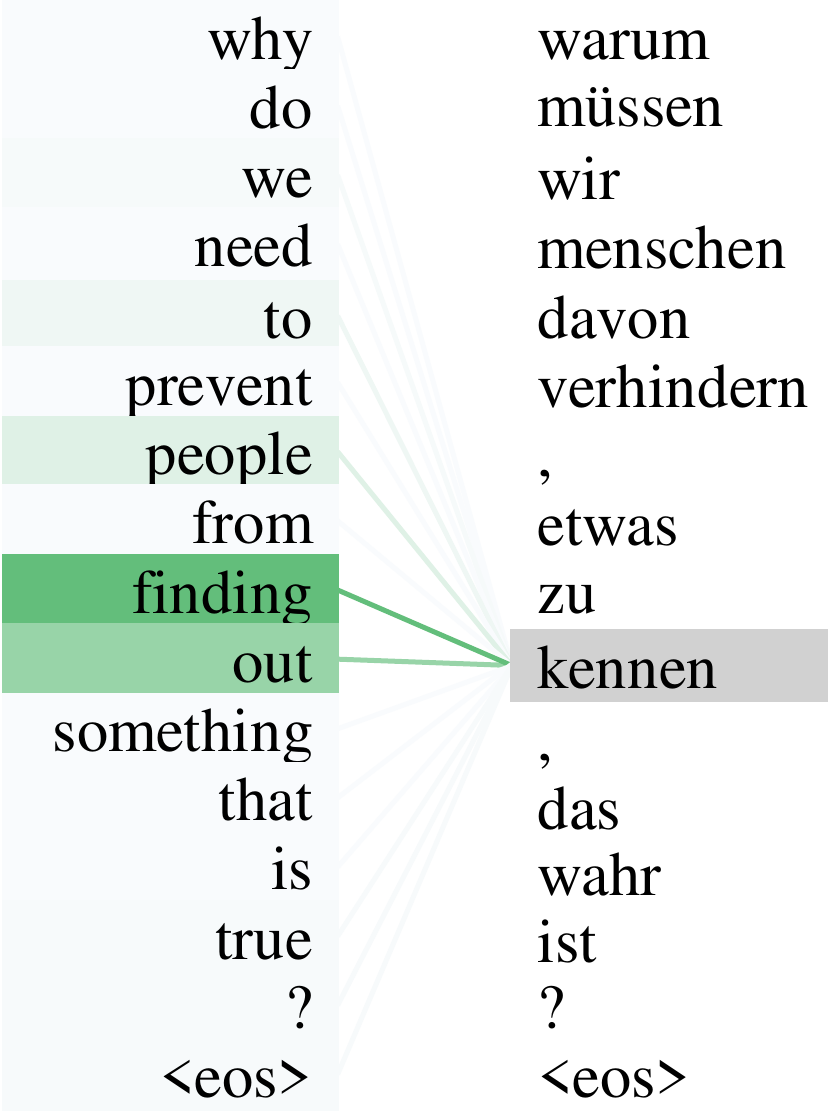}}
	\caption{\label{fig:visualization-detail}Visualization of the attention weights from the top-most layer of the decoder for both shallow and deep \textsc{Msc} nets.}
\end{figure}

\section{Visualization of Attention Weights}
\label{Appendix-C}
Visualization of attention weights for two cases that are illustrated in Figure~\ref{fig:visualization-detail}:
\begin{itemize}
	\item Case 1
	
	\textbf{Source:} \texttt{sometimes those ideas get ahead of our thinking in ways that are important.}
	
	\textbf{Reference:} \texttt{manchmal eilen diese ideen unserem denken voraus, auf ganz wichtige art und weise.}
	
	\textbf{Baseline:} \texttt{manchmal kommen diese ideen unsere denkweise voraus, die wichtig sind.}
	
	\textbf{Deep \textsc{Msc}:} \texttt{manchmal eilen diese ideen unseren denken voraus, in einer sehr wichtigen weise.}
	
	\item Case 2
	
	\textbf{Source:} \texttt{why do we need to prevent people from finding out something that is true?}
	
	\textbf{Reference:} \texttt{warum m\"ussen wir verhindern, dass menschen kenntnis von etwas erlangen, wenn es wahr ist?}
	
	\textbf{Baseline:} \texttt{warum m\"ussen wir verhindern, dass menschen etwas finden, das wahr ist?}
	
	\textbf{Deep \textsc{Msc}:} \texttt{warum m\"ussen wir menschen davon verhindern, etwas zu kennen, das wahr ist?}
\end{itemize}